%% file: root.tex
\documentclass[letterpaper, 10 pt, conference]{ieeeconf-modified}  % Comment this line out if you need a4paper

\IEEEoverridecommandlockouts                              % This command is only needed if 
\usepackage[numbers]{natbib}
\usepackage{graphicx}
\graphicspath{ {./figures} }
\usepackage{siunitx}
\usepackage{algorithm}
\usepackage{algorithmic}
\usepackage{amsmath} % assumes amsmath package installed
\usepackage{amssymb}  % assumes amsmath package installed
\usepackage{textcomp}
\usepackage{hyperref}
\usepackage[capitalize]{cleveref}
\usepackage{booktabs}
\usepackage[caption=false]{subfig}
\usepackage{xcolor}
\usepackage{mathtools}

\usepackage{etoolbox}
\makeatletter
\patchcmd{\@makecaption}
  {\scshape}
  {}
  {}
  {}
\makeatletter
\patchcmd{\@makecaption}
  {\\}
  {.\ }
  {}
  {}
\makeatother

\crefname{equation}{}{}
\definecolor{revcolor}{HTML}{636EFA}
\definecolor{todocolor}{HTML}{EF553B}
\definecolor{commentcolor}{HTML}{00CC96}
\newcommand{\rev}[1]{\textcolor{black}{#1}}
\newcommand{\rem}[1]{\iffalse\textcolor{gray}{\sout{#1}}\fi}
\newcommand{\todo}[1]{\iffalse\textcolor{todocolor}{\textbf{Todo:}\begin{itemize}#1\end{itemize}}\fi}

\newcommand{\mycomment}[1]{}

\title{\LARGE \bf Shape Completion with Prediction of Uncertain Regions}
\author{
    Matthias Humt$^{1,2}$, Dominik Winkelbauer$^{1,2}$, and Ulrich Hillenbrand$^{1}$% <-this % stops a space
    \thanks{$^1$ The author is with the DLR Institute of Robotics and Mechatronics.}
    \thanks{
        $^2$ The author is with the Technical University of Munich (TUM). \newline 
        Contact: \tt\footnotesize{matthias.humt@dlr.de}
    }
}

\begin{document}

\maketitle
\thispagestyle{empty}
\pagestyle{empty}

%%%%%%%%%%%%%%%%%%%%%%%%%%%%%%%%%%%%%%%%%%%%%%%%%%%%%%%%%%%%%%%%%%%%%%%%%%%%%%%%

\input{chapters/abstract.tex}

%%%%%%%%%%%%%%%%%%%%%%%%%%%%%%%%%%%%%%%%%%%%%%%%%%%%%%%%%%%%%%%%%%%%%%%%%%%%%%%%

\input{chapters/introduction}
\input{chapters/relatedwork}
\input{chapters/method}
\input{chapters/experiments}
\input{chapters/conclusion}

%%%%%%%%%%%%%%%%%%%%%%%%%%%%%%%%%%%%%%%%%%%%%%%%%%%%%%%%%%%%%%%%%%%%%%%%%%%%%%%%

\footnotesize
\bibliographystyle{IEEEtranN-modified}
\bibliography{IEEEabrv, bibliography}

%%%%%%%%%%%%%%%%%%%%%%%%%%%%%%%%%%%%%%%%%%%%%%%%%%%%%%%%%%%%%%%%%%%%%%%%%%%%%%%%

\end{document}

%% file: chapters/abstract.tex
\begin{abstract}
Shape completion, i.e., predicting the complete geometry of an object from a partial observation, is highly relevant for several downstream tasks, most notably robotic manipulation.
When basing planning or prediction of real grasps on object shape reconstruction, an indication of severe geometric uncertainty is indispensable.
In particular, there can be an irreducible uncertainty in extended regions about the presence of entire object parts when given ambiguous object views. To treat this important case, we propose two novel methods for predicting such uncertain regions as straightforward extensions of any method for predicting local spatial occupancy, one through postprocessing occupancy scores, the other through direct prediction of an uncertainty indicator. We compare these methods together with two known approaches to probabilistic shape completion. Moreover, we generate a dataset, derived from ShapeNet~\cite{Chang2015ShapeNetAI}, of realistically rendered depth images of object views with ground-truth annotations for the uncertain regions. We train on this dataset and test each method in shape completion and prediction of uncertain regions for known and novel object instances and on synthetic and real data. While direct uncertainty prediction is by far the most accurate in the segmentation of uncertain regions, both novel methods outperform the two baselines in shape completion and uncertain region prediction, and avoiding the predicted uncertain regions increases the quality of grasps for all tested methods.
\footnotesize{Web:~\url{https://github.com/DLR-RM/shape-completion}}
\end{abstract}

%% file: chapters/introduction.tex
\section{Introduction}

Object grasping and manipulation is the primary task of many robotic systems. As a challenging example, consider an assistance robot operating in human living environments with various objects (household objects, tools) that cannot all be known in advance. A geometric reconstruction would be a good starting point for planning all kinds of interaction. However, acquiring a full view of an object is often not practical or even impossible due to unavoidable occlusions by other objects.

The present work concerns a single view of an unknown object instance of a known category, which, in practice, may be known from an object detector. We investigate shape completion and its adequacy for predicting stable grasps on the object. This has been the subject of study in several works~\cite{Varley2016ShapeCE,Merwe2019LearningC3,Chen2022ImprovingOG}. Unlike the previous works, however, we focus on treating significant uncertainties that arise for objects with partial symmetry. Specifically, there can be identical views for objects with partial symmetry for various object poses as observed in~\cite{Saund2020DiversePS}. Thus a unique object pose is not implied for such views, and the position of occluded object parts may therefore vary within a wide spatial range.

\begin{figure}[t!]
\centering
\includegraphics[width=\linewidth]{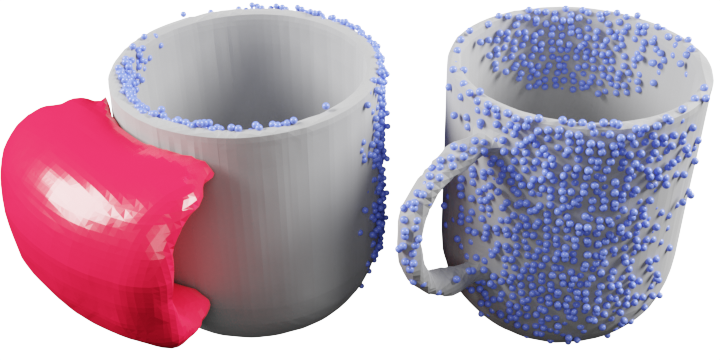}
\caption{Shape completion of a mug. Left: If the handle is occluded from the camera view, we predict the uncertain region \rem{(red)}\rev{(\textcolor{red}{\textbf{red}})} that contains the handle, resulting from pose ambiguity. The mug is reconstructed\rem{ (grey)} in the region not affected by pose ambiguity \rev{(\textcolor{gray}{\textbf{gray}})}. Right: If any part\rem{s} of the handle \rem{are}\rev{is} visible, also the handle is reconstructed, and no uncertain region is predicted.}
\label{fig:teaser}
\vspace{-1em}
\end{figure}

Examples of such objects from the household domain are mugs, jugs, and pitchers with partial rotational symmetry and a handle as the symmetry breaker.\rem{In situations when the handle is not represented in the input data due to \mbox{(self-)occlusion} or because of sensory limitations, the resulting pose uncertainty has a significant spatial extend.} In fact, an extended region may contain an occluded handle and should hence be avoided by the robot gripper; see \cref{fig:teaser} for an example. Note that this kind of uncertainty exists even for a fully known object shape or a perfect shape predictor that would provide an accurate reconstruction from another viewpoint. This uncertainty arises from the object shape and the given viewpoint, not from any model uncertainty of the predictor, that is, epistemic uncertainty. In this sense, it is an \textit{objective uncertainty}.

In this work, we consider the representative example of mug-like objects. We propose two different methods for predicting the uncertain spatial regions that result from pose ambiguity. These methods are used here as an extension of~\cite{Peng2020ECCV} for shape completion but can be combined with any completion method that predicts spatial occupancy scores (probabilistic or otherwise).\rem{ One of the\rev{se} methods \rem{for uncertain region prediction is through the}\rev{employs} postprocessing of the predicted occupancy scores. The other \rem{method}\rev{one} introduces \rem{an extra label for}\rev{explicit labeling of} the uncertain regions. The prediction is \rem{then}\rev{thus} converted into a \rem{trinary}\rev{\textit{trinary}} classification problem: free\rem{ space} vs.\ occupied\rem{ space} vs.\ uncertain space. Note that the uncertain regions have a ground-truth extension, just as the occupied space, which merits the representation with its own label.}

We conduct an extensive set of experiments on known and novel object instances and both on synthetic and real data. We compare our approaches to two existing methods for probabilistic shape completion that, after little adaptation, lend themselves to predicting viewpoint-induced uncertain regions~\cite{Lundell2019RobustGP,Saund2020DiversePS}. Our proposed methods outperform the two baselines on a range of metrics for occupied and uncertain region prediction and improve grasp quality by avoiding the predicted uncertain regions.

\rem{We conduct three kinds of experiments\rev{\ for each method and baseline}: (i) Shape completion and \rem{uncertainty}\rev{uncertain region} prediction \rem{on}\rev{from} point clouds rendered from ShapeNet~\cite{Chang2015ShapeNetAI} where test objects are randomly rescaled versions of training objects\rem{ and} from novel viewpoints; (ii) same as \rem{before}\rev{(i)} but with test objects not contained in the training set, hence novel instances; (iii) \rem{Shape completion and uncertainty prediction trained as before}\rev{same as (ii)} but with test objects from the BOP challenge~\cite{Hodan2018BOPBF, Hodan2020BOPC2}, \rem{hence demonstration of}\rev{demonstrating generalization from simulation to the real world} (sim2real)\rem{ transfer} on novel instances. We run grasp prediction on the test objects for all these cases as a downstream task. The advantage of predicting and adhering to uncertain regions is evident from the results of each experiment. Moreover, we find that the trinary prediction of uncertain regions is more accurate than those from the binary occupancy prediction\rev{\ while both outperform the two baselines}.}

We make the following contributions.
\begin{itemize}
\item Introduction of two novel methods for predicting uncertain regions in shape completion as extensions of any predictor of spatial occupancy scores.
\item Validation and comparison of these new methods and two baselines on the task of uncertainty prediction in an extensive set of experiments, showing superior performance of the new methods.
\item In these experiments, demonstration of the utility of the predicted uncertain regions as a constraint for high-quality grasps prediction.
\item A dataset of realistically rendered object views, derived from ShapeNet~\cite{Chang2015ShapeNetAI}, including ground-truth annotation of the corresponding uncertain regions. %\footnote{\url{https://github.com/DLR-RM/shape-completion}}.
\end{itemize}

%% file: chapters/relatedwork.tex
\section{Related Work}

\subsection{Shape Completion}

In the context of shape completion, i.e., reconstructing the entire 3D object shape from a partial 3D observation, learning architectures that can deal with unordered 3D point clouds \cite{Fan2016APS,Tchapmi2019TopNetSP,Yuan2018PCNPC} are of great interest, as this is the modality directly obtained from depth sensors. Point clouds can easily be voxelized in different ways and then processed with voxel-based methods \cite{Wu20143DSA,Dai2016ShapeCU,Lundell2019RobustGP,Saund2020DiversePS}. The advantage is that proven techniques from CNNs can be used in the ordered grid, but at the cost of high memory when high resolution is required.

% Reconstruction from RGB images \cite{Choy2016} is a related but different problem that can be incorporated using RGB-D sensors' color channels.

Apart from the input modality, the shape representation generated by a shape completion process can make a difference for downstream tasks. In particular, for the present context of robotic grasp generation, triangle surface meshes are the preferred choice. Some methods directly produce meshes as output \cite{Groueix2018APA,Liao2018DeepMC}. But the task is challenging because of the complex mesh structure with varying topology, and hence the generated meshes often need postprocessing to be usable.

More recently, the representation of shapes through an implicit function has emerged as a new paradigm in the shape completion community \cite{Mescheder2018OccupancyNL,Park2019DeepSDFLC,Peng2020ECCV,Chibane2020ImplicitFI}. Instead of predicting the shape of the object explicitly, the problem is cast as a point-wise classification problem where the object surface is implicitly defined as the decision boundary between its interior and exterior \cite{Mescheder2018OccupancyNL,Peng2020ECCV,Chibane2020ImplicitFI}. It can also be formulated as a regression problem where the task is to predict the signed distance from the surface \cite{Park2019DeepSDFLC} for each query point.

A significant advantage of the implicit representation is its ability to be queried at continuous locations, e.g. in a grid of arbitrary resolution to extract a detailed surface mesh. Further, it can be flexibly combined with encoders to match the input data type, such as a simple PointNet \cite{Qi2016PointNetDL} to generate complete meshes from partial point clouds. This is the desired combination for our use case.

Our work is further concerned with shape completion as a basis for geometrically informed robotic grasping, which has recently attracted wider attention \cite{Varley2016ShapeCE,Lundell2019RobustGP,Saund2020DiversePS,Merwe2019LearningC3,Chen2022ImprovingOG}. Contrary to previous work however, we focus on the specific problem of extended regions of objective uncertainty caused by viewpoint-dependent ambiguity in the input data.

While \cite{Lundell2019RobustGP} are also concerned with uncertainty in shape completion in robotic grasping, their focus lies on epistemic or model uncertainty. More related to our work, \cite{Saund2020DiversePS} focus on ambiguity arising from partial observations and the generation of diverse yet plausible shape completions. But they omit to extract the resulting regions of uncertainty and to directly use them as a constraint for grasp generation, which we propose to do in this work. We quantitatively compare our proposed methods to these two prior works, adapting them to the here-considered problem of predicting extended uncertain regions.

\subsection{Uncertainty Quantification in Pose Estimation}

The studied uncertainty for shape completion arises from uncertainty in the objects pose for partially symmetric objects. The quantification of pose uncertainty has also been addressed in some works on object pose estimation.

The uncertainty in an estimate can be quantified as the level of confidence in the prediction, sometimes called predictive uncertainty \cite{Tremblay2018DeepOP,Shi2020FastUQ}. A more expressive representation of pose uncertainty is predicting a probability density of poses. There has been work on modeling rotational uncertainty with parametric densities and their mixtures \cite{Prokudin2018DeepDS,Gilitschenski2020DeepOU}. The density parameters and mixture weights are predicted. However, to determine the spatial effect of pose uncertainty, a sufficient sample of poses would need to be drawn from the predicted densities, and for each pose, the object model needs to be transformed into the scene for mapping of the resulting occupancy. This extra effort may easily prevent real-time performance.

An interesting alternative to a parametric density model is through a deep neural network (DNN) that predicts the probability for a given image and every queried object orientation \cite{Murphy2021ImplicitPDF}. Another alternative is density representation through a particle filter for pose tracking \cite{Deng2019PoseRBPFAR}. However, the effort of sampling many poses and transforming object models into the scene for each of them remains.

The above approaches mainly target known objects; no uncertainty quantification under shape variations is studied. In the present work, we aim at the generalization across shapes.

%% file: chapters/method.tex
\section{Method}

This section describes the generation of realistic synthetic training data and training conditions required to generalize to real-world data. We then briefly describe the data representation and network architecture to predict complete shapes from partial inputs. Finally, a method for finding ambiguous views leading to uncertain regions in the shape completion is presented, and two methods for the \rem{predicting}\rev{prediction} of these uncertain regions are introduced.

\subsection{Realistic Synthetic Data}

Point clouds projected from depth images are the most natural representation of real-world 3D data but come with unique challenges beyond permutation equivariance and inhomogeneous density. Self-occlusion only allows for partial views of captured objects, while sensor characteristics and noise remove additional information.

While these effects are captured naturally for data collected with sensors in the real world, this data acquisition process is extremely time-consuming and does not scale to a diverse and potentially changing set of objects and scenarios.

We, therefore, resort to synthetic data generation through rendering from various viewpoints and subsequent online augmentation with noise. As sensors experience stronger artifacts on surface regions perpendicular to the viewing direction, we use the\rem{ inverse} cosine similarity with the per-point normal direction as the probability of removing points and adding additional noise.

\subsection{Realistic Training Conditions}

The camera pose relative to fixed world coordinates, i.e., the \textit{object} coordinate frame, is frequently unknown on a (mobile) robotic system. For this reason, shape completion approaches geared towards real-world application are usually trained in \textit{camera} coordinates, which can yield better generalization to unknown shapes, albeit at the cost of reduced fidelity on known examples~\cite{Shin2018PixelsVA}.

When working with a robotic system, we identify a third alternative to camera and object coordinates: the \textit{robot world} coordinates. Through the robot's kinematic chain, which is usually known, the camera pose can be projected to this world frame, removing the effects of camera rotation, thus reducing the complexity of the learning problem to object pose variations.

\subsection{Implicit Function Learning}

We \rem{start by formalizing}\rev{briefly summarize} the shape completion problem as implicit function learning \rev{following}~\cite{Mescheder2018OccupancyNL} \rev{which we make use of in our first approach and extend for the second one}. Given an observation in the form of a partial point cloud $x\in\mathcal{X}$ and an arbitrary 3D point $p\in\mathbb{R}^3$, we want to find a parametric function \mbox{$f_\theta:\mathbb{R}^3\times\mathcal{X}\rightarrow[0,1]$} where $f_\theta$ is a DNN, and its output represents the occupancy probability at $p$. The task of learning the parameters of the model, therefore, reduces to the standard binary classification problem \rem{and can be}\rev{which is} trained using the\rem{ standard} binary cross-entropy cost function,

Evaluating $f_\theta$ at points on a 3D grid, we\rem{ can} extract a mesh from the resulting occupancy grid \rev{as the approximate isosurface at threshold $\tau$ with $f_\theta(p,x)=\tau$} using \rev{the} \textit{Marching Cubes}~\cite{Lorensen1987MarchingCA} \rev{algorithm}\rem{ as the approximate isosurface at threshold $\tau$ as $f_\theta(p,x)=\tau$}. At $\tau=0.5$, the extracted isosurface represents the decision boundary between the two classes (\textit{free} and \textit{occupied}), defining the surface of the object implicitly.

\subsection{Uncertain Shape Completion}

Quantifying uncertainty in deep learning is of great interest~\cite{Gal2016UncertaintyID}, especially if the model is deployed in the real world where wrong predictions made with high certainty, i.e., overconfident predictions, can have grave consequences.

The predictive uncertainty of a model is the consequence of epistemic and aleatoric uncertainty. Dealing with objects in 3D space, additional \textit{pose uncertainty} can arise. Given an ambiguous object view, this form of uncertainty remains even under conditions of perfect knowledge regarding model parameters and object geometry: it is irreducible and objective. This effect can be observed for objects with partial symmetry (continuous or discrete) broken only from specific viewpoints.

Mugs are a prominent real-world example. When the handle is hidden from view by the mug's body, many possible orientations of the mug yield the same observation. As a result, there is an extended region that may contain the mugs handle, which we call \textit{uncertain region}. Crucially, however, this uncertainty is precisely defined by all rigid transformations $T\in\mathrm{SE}(3)$ that yield the same observation.

Similar to~\cite{Saund2020DiversePS}, we estimate this set of transformations, and thereby the region in $\mathbb{R}^3$ affected by pose uncertainty by randomly transforming the object and comparing the resulting 2D projection to the initial one. An identical or, in practice, similar view that places the handle at a different position enlarges the uncertain region.

Given this ground truth information about uncertain regions for each observation, we extend the binary classification task of implicit function learning with classes \textit{free} and \textit{occupied} by a third class \textit{uncertain}. We then train a new model \mbox{$f_\theta:\mathbb{R}^3\times\mathcal{X}\rightarrow[0,1,2]$}.

While this direct approach is intuitive and promising, it requires the costly identification of ground truth uncertain regions for model training. A second approach that does not need extra annotation relies on the per-class score to measure the uncertainty in a predicted local occupancy.

To extract regions of low certainty from the occupancy grid, we can set $\tau<0.5$. Unfortunately, each isosurface at $\tau_1$ is entirely surrounded by another with $\tau_2<\tau_1$. Thus, we need a way to differentiate between narrow uncertain regions necessarily occurring near the object's surface and more extended ones reaching farther out.

One such criterion is $\lvert\nabla_p\hat{y}\lvert$, the magnitude of the gradient of the occupancy probability at a query point $p$ with \mbox{$\hat{y}\coloneqq p(y=1\lvert x,p,\theta)$} being the predicted occupancy probability at point $p$ conditioned on the observation and the model parameters. Being able to obtain such gradients is a unique and valuable property of implicit function models~\cite{Mescheder2018OccupancyNL}.

\begin{figure}[htbp]
\vspace{-1.5em}
\centering
\subfloat[Predicted occupancy probability]{%
  \centering%
  \includegraphics[width=0.5\linewidth]{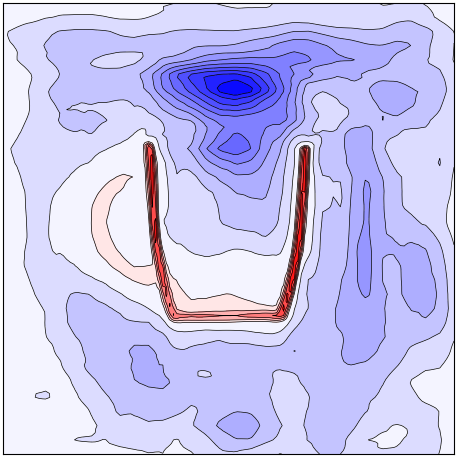}%
}
\subfloat[Gradient of (a)]{%
  \centering%
  \includegraphics[width=0.5\linewidth]{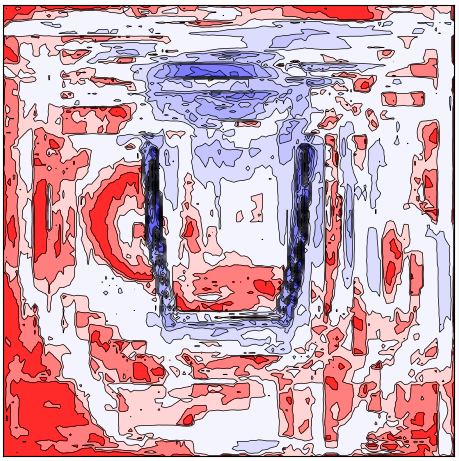}%
}
\caption{Slice of a side view from the predicted occupancy probability grid (a). Using a small lower threshold, a region possibly containing a handle appears behind the mug (light red), but the mug itself is also contained (dark red). Using the gradient of the predicted occupancy probability (b) with its average as upper threshold, this unwanted region can be discarded by only considering the intersection of regions shown in red as uncertain.}
\label{fig:contour}
\end{figure}

We expect this gradient to be large near the decision boundary defining the object's surface, whereas it should be small for points farther away. Using an upper threshold on the norm of the gradient for each point in the occupancy grid, we extract approximate uncertain regions from a trained model, as shown in \cref{fig:contour}, without relying on ground truth \textit{uncertain} labels.

While the approach with the gradient criterion does not need extra data annotation, it does require a proper adjustment of the gradient threshold, in the general setting covering all possible uncertain region sizes with their gradients.

Another possible criterion is the variance of multiple stochastic forward passes using \textit{Monte Carlo Dropout}~\cite{Gal2016DropoutAB} as employed by~\cite{Lundell2019RobustGP}.

Similarly, we are also able to extract an uncertain region from~\cite{Saund2020DiversePS} through the variance of multiple samples from the Variational Autoencoder (VAE).

For all methods, a low threshold $\tau$ extracts a larger uncertain region but often introduces artifacts in the form of small free-floating pieces of geometry. As we want to extract large continuous areas, we cluster the connected components and remove those too small.

%% file: chapters/experiments.tex
\section{Experiments}

We conduct three experiments to evaluate (i) the generalization ability to novel views of known objects, (ii) the ability to generalize to novel instances of the same class, and (iii) the generalization from synthetic to real data (sim2real). The evaluation is done regarding shape completion, prediction of uncertain regions, and its effect on the robotic task of object grasping. We compare the two newly proposed methods, i.e., the \textit{binary} with a gradient criterion and the \textit{trinary}, against each other and the two adapted baselines, i.e., \textit{dropout} and \textit{VAE}, described in the previous section.

\subsection{Datasets}

We render 200k train, 20k validation, and 40k test depth and normal maps offline using \textit{BlenderProc2}~\cite{Denninger2023BP} and then augment them online during training as shown in \cref{alg:augment} and explained in the previous section.

\begin{figure}[htbp]
\vspace{-1em}
\begin{algorithm}[H]
  \caption{Generate datum}\label{alg:augment}
  \begin{algorithmic}[1]
    \STATE Sample camera position {c} randomly in upper hemisphere
    \STATE Scale object $\mathcal{U}(0.05,0.15)$ and $\pm$20\% along z-axis
    \STATE Render depth and normal image
    \STATE Project depth, normals to point cloud $\mathcal{P}=\{(p_i,n_i)\}_{i=0}^N$
    \FOR{$p,n\in\mathcal{P}$}
      \STATE $p=p+\mathcal{N}(0,0.005)$  // Add noise
      \STATE $s=n\cdot c/\|n\|\|c\|$  // Compute cosine similarity
      \STATE $\mathcal{P}\setminus p$ if $\vert s\vert<\mathcal{U}(0,1)$  // Remove point
      \STATE $p=p+\mathcal{N}(0,0.01)$ if $\vert s\vert<\mathcal{U}(0,1)$  // Add noise
    \ENDFOR
    \STATE Transform point cloud from camera to robot world frame
    \STATE \textbf{return} augmented point cloud
  \end{algorithmic}
\end{algorithm}
\vspace{-1em}
\end{figure}

We consider two datasets for our experiments. For the novel view and novel instance experiments, we use the \textit{mugs} category of the \textit{ShapeNet}~\cite{Chang2015ShapeNetAI} dataset. \rem{There are}\rev{Of the} 214 available mugs, we select the 201 with a handle. Watertight meshes and query points are generated similarly to~\cite{Mescheder2018OccupancyNL}.
For the novel instance experiment, we split the 201 objects into training (70\%), validation (10\%), and test (20\%) sets.

The sim2real experiment is conducted on a subset of the datasets from the \textit{BOP Challenge}~\cite{Hodan2018BOPBF, Hodan2020BOPC2}\rem{. While its original intention is to capture the STATE of the art in estimating the 6D pose, the provided real\rem{-}\rev{\ }world RGB-D images with ground-truth pose annotations in camera coordinates for the visible objects as well as their meshes\rem{ exactly} match the requirements for testing shape completion.}\rev{, which provide real-world RGB-D images with ground truth per-object pose, segmentation mask and mesh.}

\rem{There are five}\rev{Five} out of\rem{ a total of} 12 datasets (\textit{Linemod (LM), HomebrewedDB (HB), YCB-Video (YCB-V), Toyota Light (TYOL)}) contain\rem{ing} a total of nine different mugs. We replace the \rem{extremely} low quality \textit{LM} mug mesh with a higher quality mesh of the featured \textit{IKEA FÄRGRIK} mug and transform the \textit{HB} mug from being half full to empty. HB provides RGB-D data from both the Primesense ($\text{HB}_{\rm pri}$) and Kinect ($\text{HB}_{\rm kin}$) sensors. We use both YCB-V sequences (48 and 55) in which the mug is present.

As the provided ground truth poses and/or camera intrinsic parameters are inaccurate, we perform a few steps of the \textit{Iterative Closest Point}~\cite{Besl1992AMF} algorithm between\rem{ the} projected\rev{\ real and rendered} depth maps\rem{ and the mesh} to improve \rem{its}\rev{the ground truth} pose.

We use the provided per-object segmentation masks to remove \rem{their surroundings}\rev{non-object points}. Those masks are inaccurate, however, leading to unwanted pixels at the boundaries, which we remove by scaling the convex hull of the object mesh by a small factor and only keep\rev{ing} points on the inside. We only evaluate views with an object visibility of at least $85\%$.

\subsection{Metrics}

We evaluate the performance in region segmentation, both occupied and uncertain, using the standard volumetric metrics \textit{IoU}, \textit{Precision}, \textit{Recall}, \textit{F1-Score}. We further build three grasping-related metrics from the confusion matrix constituents.
\begin{align*}
    \text{GCR} & = (\text{FN}_{\rm occ} + \text{FN}_{\rm unc}) / (\text{TP}_{\rm occ} + \text{FN}_{\rm occ} + \text{TP}_{\rm unc} + \text{FN}_{\rm unc}) \\
    \text{GMR} & = \text{FP}_{\rm occ} / (\text{FP}_{\rm occ} + \text{TP}_{\rm occ}) \\
    \text{GER} & = \lvert\mathcal{FP}_{\rm occ}\cup\mathcal{FP}_{\rm unc}\rvert / (\text{FP}_{\rm occ} + \text{TN}_{\rm occ})
\end{align*}
Here FP, FN, TP, and TN are the usual true and false counts for classifying occupied (occ) and uncertain (unc) vs.\ free regions, respectively. Confusion between occupied and uncertain regions does not affect these metrics. $\mathcal{FP}$ denotes the set of false positives to prevent accounting twice for the same location. The \textit{Grasp Collision Risk} (GCR) is a straightforward measure of the probability of unexpected collision when closing the hand around a planned grasp region; \textit{Grasp Miss Risk} (GMR) is for the opposite case when closing the hand, i.e., an expected contact is not made; \textit{Grasp Exclusion Risk} (GER) quantifies the risk of rejecting feasible grasp candidates because they are blocked by false prediction of occupied or uncertain regions. All volume measures are quantified by counting uniform point samples inside the considered volumes.

Intuitively, precision is the fraction of correctly predicted occupied space, and recall the fraction of actual occupied space found by the model.

In addition to the volumetric quantities, we also include the \textit{Chamfer-L1} distance (CD)~\cite{Mescheder2018OccupancyNL} to measure shape difference and displacement between the surface meshes of the ground-truth and predicted occupied and uncertain regions, respectively. The length scale is set by normalizing object size to fit in a unit cube.

To measure the effect of uncertain region prediction on object grasping, we use the network from~\cite{Winkelbauer2022ATL} to predict $1024$ grasps for a humanoid four-finger hand (DLR-Hand II~\cite{Butterfass2001DLR-Hand-II}) per object from its completion. or the uncertainty-aware variant we use the predicted uncertain region as a filter, removing grasps that intersect with it. We evaluate the grasps' quality using the \textit{Improved Epsilon Quality} (IEQ)~\cite{Winkelbauer2022ATL} metric, which measures the minimal external force applied to the object (ground truth mesh) that would break the grasp where the highest IEQ in the sample is the achieved grasp quality.

\subsection{Network, Training \& Hyperparameters}

We \rem{choose}\rev{apply our proposed methods to} the \textit{Convolutional Occupancy Network} (ConvONet)~\cite{Peng2020ECCV}\rem{ as our network architecture for all experiments}, \rem{having shown}\rev{which strikes a} good \rem{performance while being lightweight}\rev{balance between performance and size}.\rem{ The latter \rem{reduces}\rev{can help to reduce} overfitting and decreases training and inference time; an important criterion in robotics applications.}

We implement all methods in \textit{Python} and \textit{PyTorch}~\cite{Paszke2019PyTorchAI}, building on the code base of~\cite{Mescheder2018OccupancyNL,Peng2020ECCV}\rev{\ for ConvONet and adapt~\cite{Lundell2019RobustGP} for our first baseline. For our second baseline, we convert the \textit{TensorFlow}~\cite{Tensorflow2015} implementation of~\cite{Saund2020DiversePS} into PyTorch}. We use the \textit{Adam}~\cite{Kingma2014AdamAM} optimizer with default parameters\rem{ except for the learning rate, which we set to $10^{-4}$} to train the models until convergence or until their validation error starts to increase.

\begin{figure}[htbp]
\includegraphics[width=\linewidth]{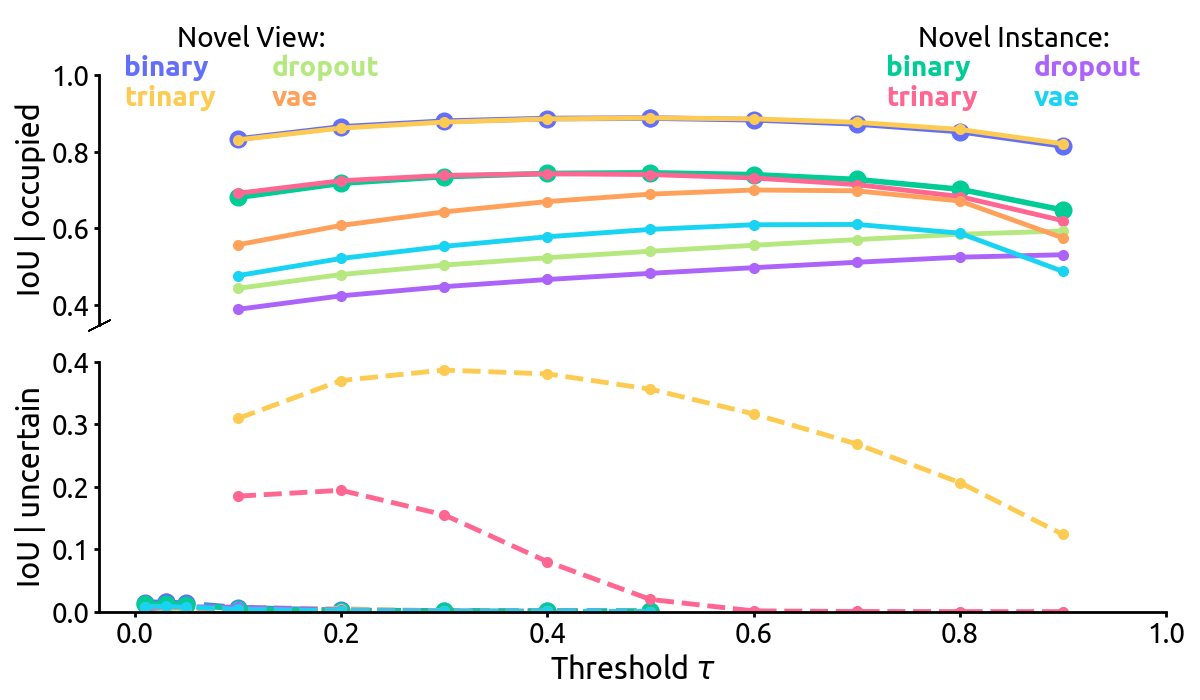}
\caption{Effect of varying the threshold parameter $\tau$ on IoU for occupied and uncertain region predictions.}
\label{fig:hyper}
\end{figure}

When extracting occupied and uncertain regions from a binary occupancy grid, two hyperparameters must be chosen: the threshold $\tau$ on the occupancy score \rev{ $\hat y$} deciding whether a point lies on the interior of the shape and another threshold $\tau_u<\tau$ determining whether a point location is regarded as uncertain. The canonical evaluation in binary classification setting requires $\tau=0.5$, but we can also view it as a hyperparameter to be optimized on the validation set for maximum IoU~\cite{Mescheder2018OccupancyNL}, shown in Fig. \ref{fig:hyper}. This can be seen as a form of post hoc calibration of the predicted probabilities similar to temperature scaling~\cite{Guo2017Calibration}.

In the trinary case, the class prediction is usually the \textit{argmax} of the predicted class scores. As in the binary case, however, we regard it as a hyperparameter and again optimize it on the validation set for maximum IoU to mitigate potential overconfidence of the DNN~\cite{Guo2017Calibration}.

\rev{One additional hyperparameter must be chosen for the \textit{binary}, \textit{dropout} and \textit{VAE} methods. For a location with $\hat y<\tau_u$  to be considered part of the uncertain region, we threshold the magnitude of the gradient to be below average for the \textit{binary} method and the variance of the samples from the \textit{dropout} and \textit{VAE} methods to be above average.}

\input{tables/novel_view.tex}
\input{tables/novel_instance.tex}
\input{tables/sim2real.tex}

\subsection{Novel View Generalization}

\cref{table:novel_view} shows the quantitative evaluation of all methods on the test set. The implicit function models (\textit{binary} and \textit{trinary}) perform similarly in predicting the occupied space and significantly outperform the voxel-based baselines. The \textit{dropout} model performs worse than the \textit{VAE}, which uses a higher-resolution voxel grid. Predicting the uncertain region is a more challenging task where the \textit{trinary} model outperforms all other methods by a large margin, followed by the \textit{binary} model.

The IEQ increases substantially across all methods when the predicted uncertain region is considered with the \textit{trinary} model taking the lead. In this case, the GER also increases, but the \textit{trinary} model is affected the least due to its ability to accurately predict the uncertain region. \rem{As is evident by the uncertain region recall, the \textit{binary} model more severely underestimates its extent compared to the \textit{trinary} model, leading--in conjunction with a high occupied region recall--to the lowest GCR.}\rev{However, the \textit{binary} model achieves the lowest GCR and GMR.} Again, both baselines are outperformed across the grasp metrics, except the \textit{dropout} model beating the \textit{binary} in IEQ.

\subsection{Novel Instance Generalization}

\cref{table:novel_instance} shows the evaluation of the \textit{novel instance} models on the held-out objects from the test set. The results from the novel view experiment are consolidated, mirroring the similar performance\rev{\ between \textit{binary} and \textit{trinary}} on the occupied region and superiority of\rem{ the} \textit{trinary}\rem{ model} in predicting the uncertain region. Nonetheless, the \textit{binary} model takes the lead in IEQ by a small margin over the \textit{trinary}. Again, both baselines are not competitive. As expected, the general performance is lower than in the novel view case as the learning task is more challenging.

\begin{figure}[htbp]
\vspace{-1.5em}
\centering
\subfloat{
  \centering
  \includegraphics[width=0.3\linewidth]{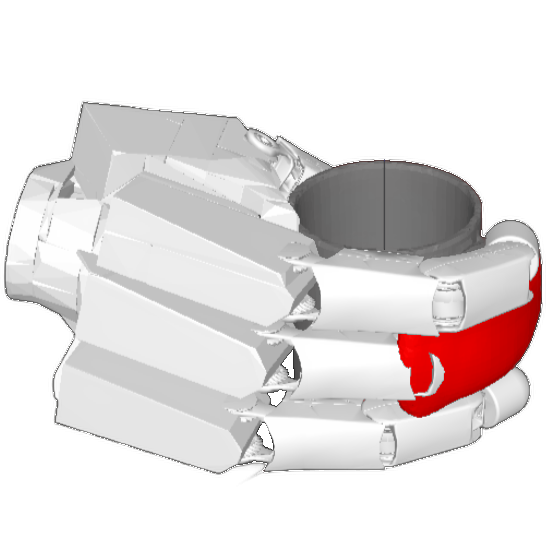}
}
\hspace{0.04\textwidth}
\subfloat{
  \centering
  \includegraphics[width=0.23\linewidth]{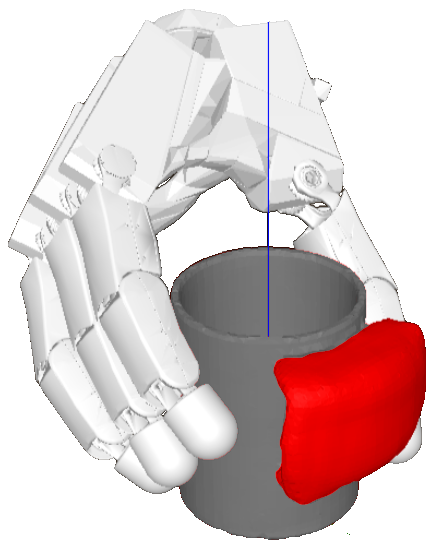}
}
\caption{Grasping the mug with an occluded handle without filtering using the predicted uncertain region leads to collision with the handle (left). Discarding grasps that collide with the uncertain region avoids collision and thus improves the grasp quality (right).}
\label{fig:grasp}
\vspace{-1.5em}
\end{figure}

\subsection{Sim2Real Generalization}

\cref{table:sim2real} shows the performance of the \textit{novel instance} models on real data and thus their sim2real capabilities. The average performance of all methods across all metrics decreases noticeably. Many effects observed for real RGB-D sensors are hard to simulate, such as the effect of surface color, opacity, and reflectance on the depth data, leading to a large sim2real gap. Visual inspection of the depth data reveals that large parts of the objects are frequently missing although visible in the RGB image, removing in particular thin structures like the mugs' handle. The effects on shape completion and prediction of uncertain regions are twofold: (i) non-occluded handles are often not reconstructed; (ii) false uncertain regions are often predicted on the back side of mugs. This mainly explains the large drop in performance of all methods.

\begin{figure}[htbp]
\centering
\begin{minipage}{.15\linewidth}
  \centering
  \includegraphics[width=\linewidth]{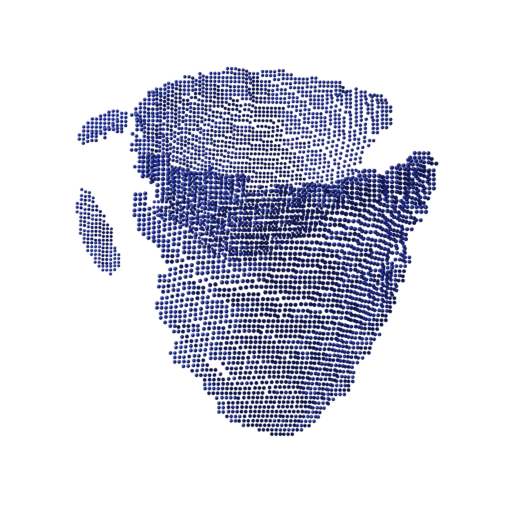}
\end{minipage}
\begin{minipage}{.15\linewidth}
  \centering
  \includegraphics[width=\linewidth]{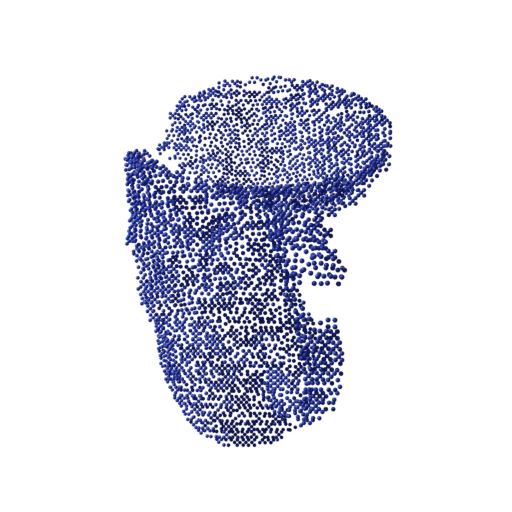}
\end{minipage}
\begin{minipage}{.15\linewidth}
  \centering
  \includegraphics[width=\linewidth]{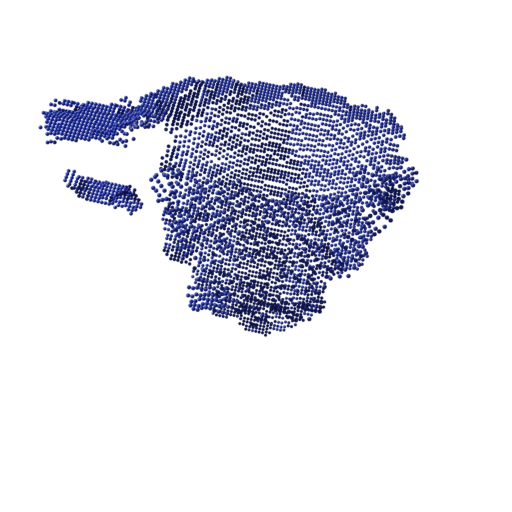}
\end{minipage}
\begin{minipage}{.15\linewidth}
  \centering
  \includegraphics[width=\linewidth]{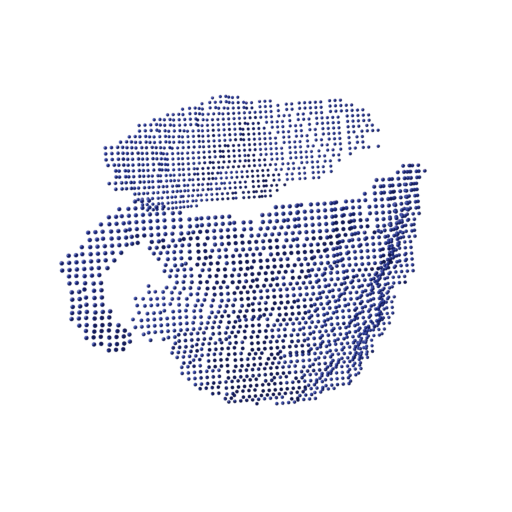}
\end{minipage}
\begin{minipage}{.15\linewidth}
  \centering
  \includegraphics[width=\linewidth]{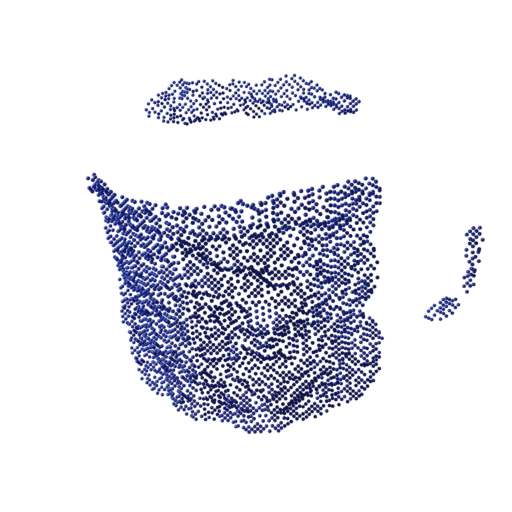}
\end{minipage}
\begin{minipage}{.15\linewidth}
  \centering
  \includegraphics[width=\linewidth]{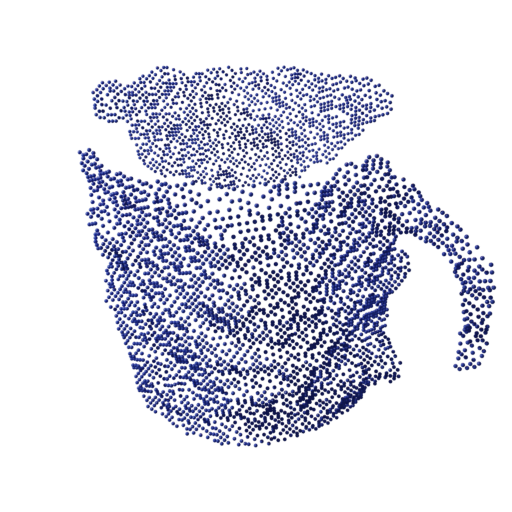}
\end{minipage}
\begin{minipage}{.15\linewidth}
  \centering
  \includegraphics[width=\linewidth]{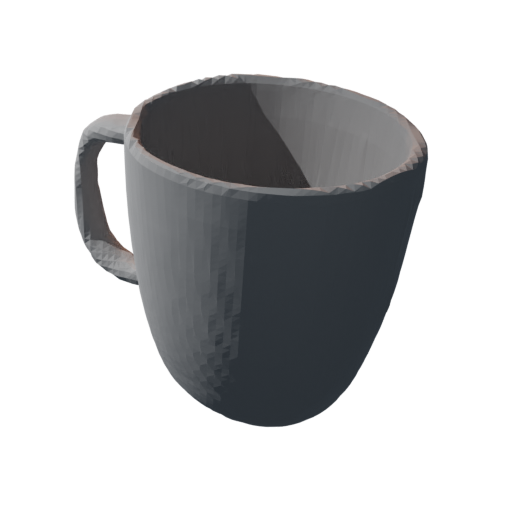}
\end{minipage}
\begin{minipage}{.15\linewidth}
  \centering
  \includegraphics[width=\linewidth]{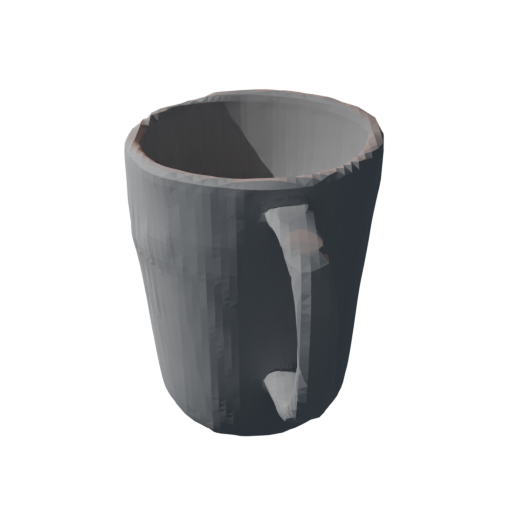}
\end{minipage}
\begin{minipage}{.15\linewidth}
  \centering
  \includegraphics[width=\linewidth]{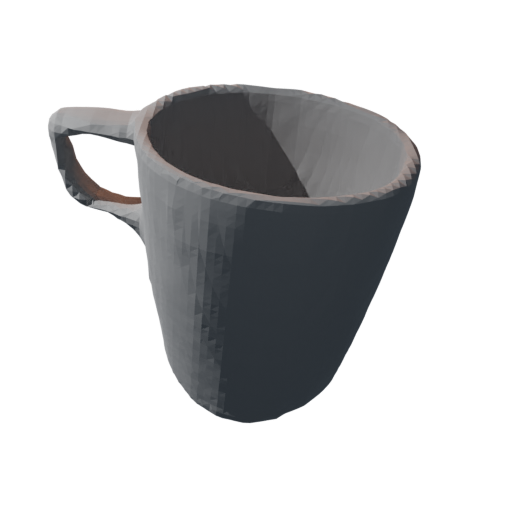}
\end{minipage}
\begin{minipage}{.15\linewidth}
  \centering
  \includegraphics[width=\linewidth]{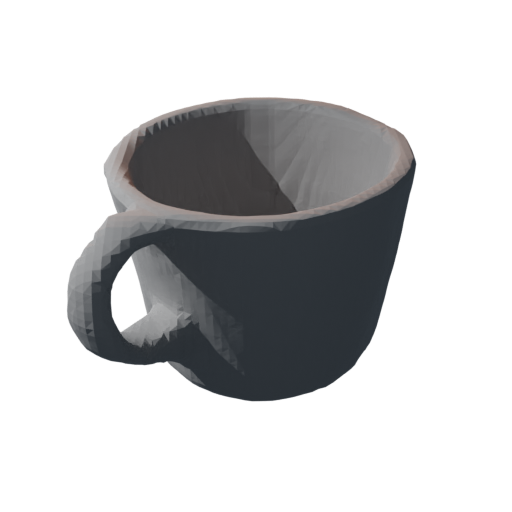}
\end{minipage}
\begin{minipage}{.15\linewidth}
  \centering
  \includegraphics[width=\linewidth]{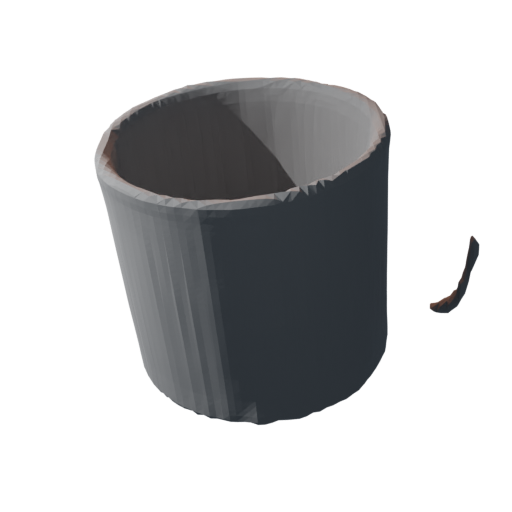}
\end{minipage}
\begin{minipage}{.15\linewidth}
  \centering
  \includegraphics[width=\linewidth]{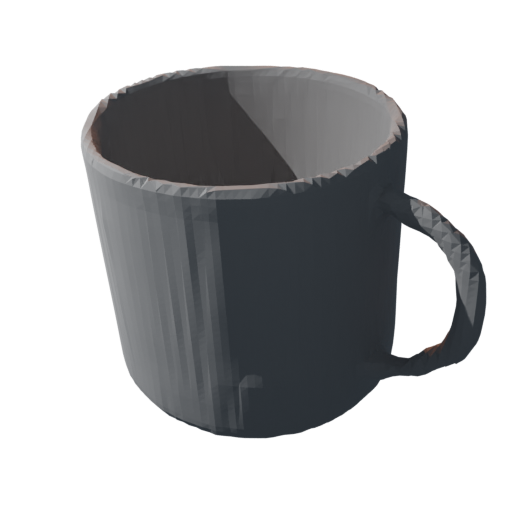}
\end{minipage}
\begin{minipage}{.15\linewidth}
  \centering
  \includegraphics[width=\linewidth]{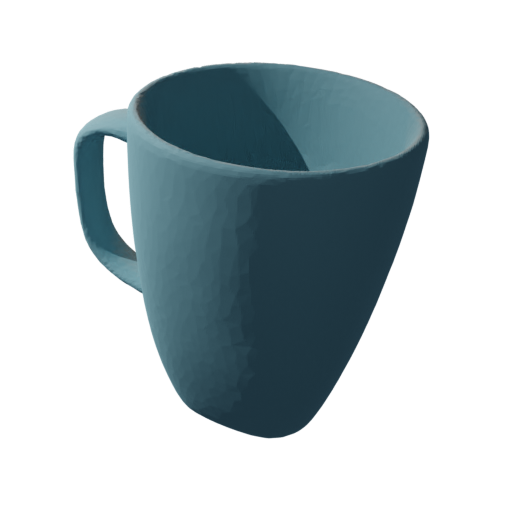}
\end{minipage}
\begin{minipage}{.15\linewidth}
  \centering
  \includegraphics[width=\linewidth]{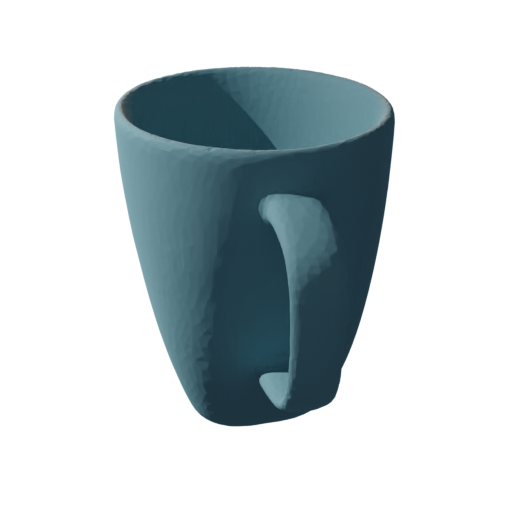}
\end{minipage}
\begin{minipage}{.15\linewidth}
  \centering
  \includegraphics[width=\linewidth]{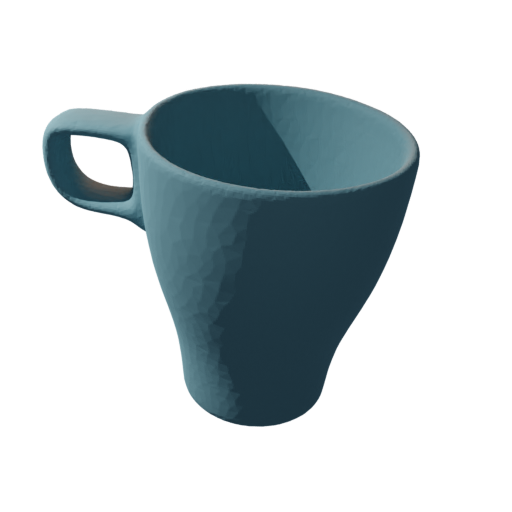}
\end{minipage}
\begin{minipage}{.15\linewidth}
  \centering
  \includegraphics[width=\linewidth]{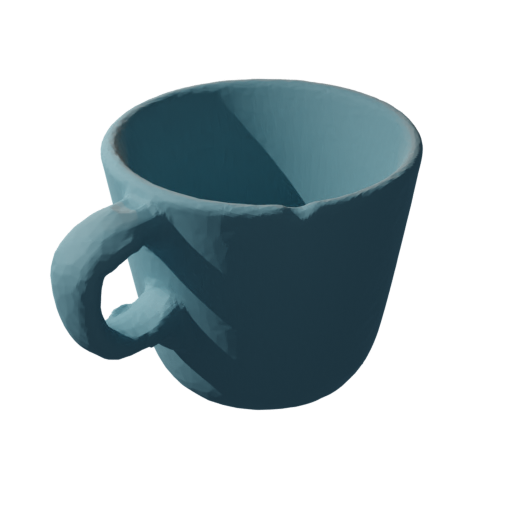}
\end{minipage}
\begin{minipage}{.15\linewidth}
  \centering
  \includegraphics[width=\linewidth]{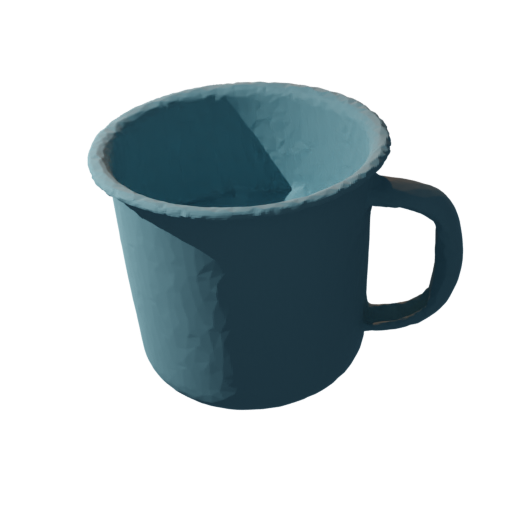}
\end{minipage}
\begin{minipage}{.15\linewidth}
  \centering
  \includegraphics[width=\linewidth]{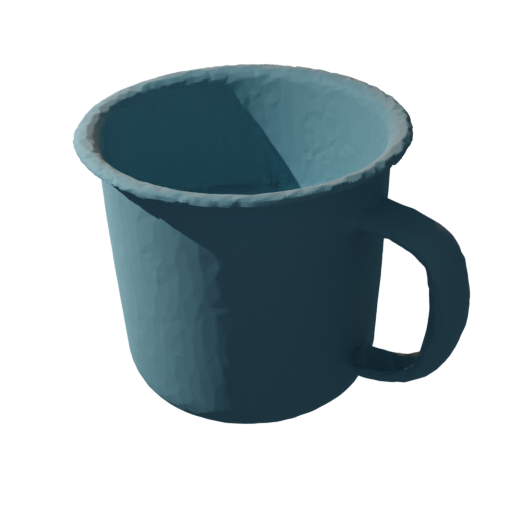}
\end{minipage}
\caption{Qualitative sim2real results. From top to bottom: Input point cloud, predicted mesh, ground-truth mesh. From left to right: test datasets $\text{HB}_{\rm pri}$, $\text{HB}_{\rm kin}$, LM, TYOL, $\text{YCBV}_{48}$, $\text{YCBV}_{55}$.}
\label{fig:sim2real}
\vspace{-0.5em}
\end{figure}

Some reconstruction errors are especially accentuated by the high sensitivity of volumetric measures in the case of low-volume objects like mugs.\rem{\footnote{Note that the mug object volume does not include the contained free space. Thus even a very well-reconstructed mug will be awarded very low IoU if it is only slightly too small/big or slightly off-center.}} However, we can see good CD values for the datasets with a better depth quality, i.e., \ $\text{HB}_{\rm pri}$ and TYOL. Moreover, qualitative results shown in \cref{fig:sim2real} reveal usable completions.

The number of images with an occluded handle having a ground-truth uncertain region is smaller than with a non-occluded handle missing from the depth image, hence with an erroneous uncertain region predicted on the back side. Consequently, a quantitative evaluation of the segmentation of uncertain regions is not sensible and therefore omitted.

Despite the--frequently large--discrepancy between the predicted and the actual uncertain regions, e.g. in most cases where handle data is missing in the depth image, we can still see an average positive effect in the IEQ values when considering predicted uncertainty in grasp planning.

\subsection{Discussion}

From \cref{fig:hyper} and \cref{table:novel_view,table:novel_instance} it is clear that the \textit{trinary} method yields by far the highest accuracy in segmenting the uncertain regions. Nonetheless, the grasp qualities achieved when considering the predicted regions do not generally stand out, which might seem odd at first sight. However, the reason for it can be established as follows.

The robot hand for which we predict the grasp (DLR-Hand II~\cite{Butterfass2001DLR-Hand-II}) is large compared to the mug objects. Hence even if the prediction of the uncertain region has low precision, a moderate recall is sufficient to likely cause an intersection with the large fingers in that region, and thus the grasp is rejected as desired. However, the low precision is still reflected in a higher GER value. Indeed, for the \textit{dropout} and \textit{VAE} methods, there were cases where all predicted grasps were blocked by the predicted uncertain region, including too many false positives, resulting in grasping failure.

In applications with specific grasp requirements, e.g.\ regarding the approach direction or the object parts to be touched, the advantage of a more accurate prediction of uncertain regions is likely to make a significant difference. In addition to grasping, an accurate prediction of viewpoint-induced uncertainty regions is also relevant in many mobile robot applications.

%% file: tables/novel_view.tex
\renewcommand{\tabcolsep}{2pt}
%\begin{table}[h]
\begin{table}[h!]
\caption{\textbf{Novel Views:} Quantitative shape completion results. All segmentation metrics are shown separately for the predicted occupied and uncertain regions. All grasp-related metrics are shown for the cases where the predicted uncertainty is ignored and considered. For metrics with $\uparrow$ higher is better, for those with $\downarrow$ lower is better. All volumetric measures are in \%, CD is $\times 100$. Best values are in \textbf{bold}.}
\vspace{-0.5em}
\label{table:novel_view}
\centering
\begin{tabular}{l|rrrr|rrrr}
\toprule
           &   \textbf{trinary} &   \textbf{binary} &   \textbf{dropout} &    \textbf{vae} &   \textbf{trinary} &   \textbf{binary} &   \textbf{dropout} &  \textbf{vae} \\
\midrule
\multicolumn{1}{l}{} & \multicolumn{4}{c}{\textit{occupied region}} & \multicolumn{4}{c}{\textit{uncertain region}} \\
\midrule
 IoU $\uparrow$      & 78.83     & \bf79.00  &     54.92 & 63.07 & \bf31.05   & 9.32     &      4.92 &  4.86 \\
 F1 $\uparrow$       & 86.08     & \bf86.29  &     69.27 & 75.38 & \bf49.53   & 31.76    &     19.38 & 18.72 \\
 Prec. $\uparrow$    & 84.97     & \bf85.59  &     64.66 & 72.32 & \bf44.63   & 13.05    &      7.19 &  7.85 \\
 Rec. $\uparrow$     & \bf90.92   & 90.60    &     79.50 & 82.72 & \bf49.51   & 45.46    &     26.16 & 23.58 \\
 CD $\downarrow$     & 2.85      & \bf2.77   &      5.24 &  4.50 & \bf15.77   & 22.86    &     21.81 & 23.56 \\
\midrule
\multicolumn{1}{l}{} & \multicolumn{4}{c}{\textit{uncertainty ignored}} & \multicolumn{4}{c}{\textit{uncertainty considered}} \\
\midrule
 IEQ $\uparrow$      & \bf1.79    & 1.64     &      1.66 &  1.66 & \bf2.25    & 2.05     &      2.20 &   1.73 \\
 GER $\downarrow$    & 2.48      & \bf2.31   &      4.89 &  3.78 & \bf2.83    & 4.73     &      7.93 &   6.27 \\
 GCR $\downarrow$    & \bf18.07   & 18.37    &     28.25 & 24.78 & 14.49     & \bf13.60  &     25.63 &  22.37 \\
 GMR $\downarrow$    & 18.06     & \bf17.07  &     33.16 & 26.77 & 18.06     & \bf17.07  &     33.16 &  26.77  \\
\bottomrule
\end{tabular}
\vspace{-0.5em}
\end{table}

%% file: tables/novel_instance.tex
\renewcommand{\tabcolsep}{2pt}
%\begin{table}[h]
\begin{table}[h!]
\caption{\textbf{Novel Instances:} Metrics as for \textit{Novel Views}; cf.\ \cref{table:novel_view}.}
\vspace{-0.5em}
\label{table:novel_instance}
\centering
\begin{tabular}{l|rrrr|rrrr}
\toprule
           &   \textbf{trinary} &   \textbf{binary} &   \textbf{dropout} &    \textbf{vae} &   \textbf{trinary} &   \textbf{binary} &   \textbf{dropout} &  \textbf{vae} \\
\midrule
\multicolumn{1}{l}{} & \multicolumn{4}{c}{\textit{occupied region}} & \multicolumn{4}{c}{\textit{uncertain region}} \\
\midrule
 IoU $\uparrow$      & 69.24     & \bf69.94  &     52.52 & 58.02 & \bf19.36   & 6.56     &      4.93 &  5.46 \\
 F1 $\uparrow$       & 80.13     & \bf80.74  &     67.63 & 71.55 & \bf35.51   & 23.11    &     18.89 & 20.07 \\
 Prec. $\uparrow$    & 81.63     & \bf83.68  &     65.25 & 72.43 & \bf32.04   & 7.72     &      7.08 &  7.70 \\
 Rec. $\uparrow$     & \bf82.91   & 81.86    &     75.19 & 75.20 & 38.20     & \bf52.07  &     27.62 & 28.19 \\
 CD $\downarrow$     & 3.69      & \bf3.64   &      5.56 &  5.11 & \bf17.05   & 25.32    &     21.95 & 25.64 \\
 \midrule
 \multicolumn{1}{l}{} & \multicolumn{4}{c}{\textit{uncertainty ignored}} & \multicolumn{4}{c}{\textit{uncertainty considered}} \\
 \midrule
 IEQ $\uparrow$      & 1.86      & \bf1.97   &      1.62 &  1.71 & 2.30      & \bf2.35   &      2.07 &   1.94 \\
 GER $\downarrow$    & 2.57      & \bf2.05   &      4.48 &  3.58 & \bf2.77    & 7.12     &      8.05 &   6.92 \\
 GCR $\downarrow$    & \bf26.70   & 27.55    &     33.34 & 31.25 & 25.79     & \bf21.98  &     30.64 &  28.33 \\
 GMR $\downarrow$    & 19.85     & \bf16.60  &     32.14 & 26.87 & 19.85     & \bf16.60  &     32.14 &  26.87  \\
\bottomrule
\end{tabular}
\vspace{-1.5em}
\end{table}

%% file: tables/sim2real.tex
\renewcommand{\tabcolsep}{3pt}
\begin{table*}[t!]
\vspace{0.6em}
\caption{\textbf{Sim2Real:} The segmentation metrics are shown for the predicted occupied region. IEQ is shown when predicted uncertainty is ignored and considered. All volumetric measures are in \%, CD is $\times 100$. Best mean values across the different test datasets are in \textbf{bold}.}
\vspace{-0.5em}
\label{table:sim2real}
\centering
\begin{tabular}{lrrrrrr|r|rrrrrr|r|r|r}
\toprule
\multicolumn{1}{l}{} & \multicolumn{7}{c}{\textbf{trinary}} & \multicolumn{7}{c}{\textbf{binary}} & \multicolumn{1}{c}{\textbf{dropout}} & \multicolumn{1}{c}{\textbf{vae}} \\
\midrule
                     &   $\text{HB}_{\rm pri}$ &   $\text{HB}_{\rm kin}$ &    LM &   TYOL &   $\text{YCBV}_{48}$ &   $\text{YCBV}_{55}$ &   \bf mean &   $\text{HB}_{\rm pri}$ &  $\text{HB}_{\rm kin}$ &    LM &   TYOL &   $\text{YCBV}_{48}$ &   $\text{YCBV}_{55}$ &   \bf mean  & \bf mean & \bf mean \\
\midrule
\multicolumn{17}{c}{\textit{occupied region}} \\
\midrule
 IoU $\uparrow$   &      49.26 &      32.22 & 33.83 &  39.64 &       23.08 &       19.54 & \bf32.93 &      49.51 &      30.45 & 32.70 &  39.92 &       22.80 &       20.25 & 32.60   &  29.37 & 29.04  \\
 F1 $\uparrow$    &      65.82 &      46.40 & 48.43 &  55.87 &       37.16 &       31.35 & \bf47.50 &      66.03 &      44.65 & 47.23 &  56.09 &       36.84 &       32.26 & 47.18   &  44.07 & 43.16  \\
 Prec. $\uparrow$ &      68.34 &      49.47 & 46.65 &  59.23 &       33.04 &       38.26 & 49.16   &      73.13 &      47.21 & 47.27 &  60.69 &       37.79 &       41.28 & \bf51.23 &  46.02 & 44.05  \\
 Rec. $\uparrow$  &      64.31 &      48.02 & 55.46 &  54.73 &       48.51 &       28.02 & \bf49.84 &      60.53 &      48.25 & 52.00 &  53.69 &       38.40 &       28.43 & 46.88   &  47.61 & 46.83  \\
 CD $\downarrow$  &       4.14 &      13.47 & 16.85 &   5.05 &       21.29 &       13.35 & \bf12.36 &       4.01 &      13.93 & 18.89 &   5.28 &       21.38 &       16.32 & 13.30   &  14.37 & 14.33  \\
 GMR $\downarrow$ &      32.13 &      53.68 & 53.81 &  43.00 &       69.97 &       57.44 & 51.67   &      26.62 &      56.48 & 53.04 &  41.04 &       64.00 &       54.17 & \bf49.23 &  54.73 & 55.90  \\
\midrule
\multicolumn{17}{c}{\textit{uncertainty ignored}} \\
\midrule
 IEQ $\uparrow$   &       0.85 &       1.15 &  0.72 &   1.86 &        0.24 &        0.79 & 0.93    &       1.09 &       1.40 &  1.11 &   2.07 &        0.69 &        0.65 & \bf1.17  &   1.07 & 1.10   \\
\midrule
\multicolumn{17}{c}{\textit{uncertainty considered}} \\
\midrule
 IEQ $\uparrow$   &       1.00 &       1.23 &  0.78 &   1.86 &        0.22 &        0.79 & 0.98    &       1.26 &       1.58 &  1.13 &   1.83 &        0.65 &        0.61 & 1.17    &   1.17 & \bf1.22 \\
% GMR $\downarrow$        &      26.62 &      56.48 &  53.04 &  41.04 &       64.00 &       54.17 & \bf49.23 &      32.13 &      53.68 &  53.81 &  43.00 &       69.97 &      57.44 & 51.67   &  54.73 &  55.90 \\
% GER $\downarrow$        &       4.19 &       8.50 &   4.93 &   3.32 &       10.32 &        3.03 & 5.71    &       2.77 &       5.25 &   3.52 &   2.64 &        7.82 &       2.63 & \bf4.11  &   6.80 &   6.99 \\
% GCR $\downarrow$        &      41.96 &      54.12 &  47.24 &  47.52 &       61.46 &       71.55 & 53.97   &      39.43 &      54.60 &  44.31 &  46.72 &       51.32 &       71.96 & \bf51.39 &  53.28 &  54.07 \\
\bottomrule
\end{tabular}
\vspace{-1em}
\end{table*}

%% file: chapters/conclusion.tex
\section{Conclusion}

In this work, we introduced and compared methods for shape completion with uncertain regions relevant for safety critical real-world applications. We focused on viewpoint-induced irreducible uncertainty about the positions of object parts, considering the case of mugs with handles and the task of robotic grasping. We showed the advantage of predicting and avoiding uncertain regions in extensive experiments on synthetic and real data and the superiority of two novel methods over two existing, adapted baselines. Future work will consider more object classes and grasping scenarios and further improve sim2real capabilities.